\documentclass[lettersize,journal]{IEEEtran}
\usepackage{amsmath,amsfonts}
\usepackage{algorithmic}
\usepackage{algorithm}
\usepackage{array}
\usepackage[caption=false,font=normalsize,labelfont=sf,textfont=sf]{subfig}
\usepackage{textcomp}
\usepackage{stfloats}
\usepackage{url}
\usepackage{verbatim}
\usepackage{graphicx}
\usepackage{cite}
\usepackage[dvipsnames]{xcolor}
\usepackage{multirow}
\usepackage{graphicx}
\usepackage{amsmath}
\usepackage{amssymb}
\usepackage{booktabs}
\usepackage{adjustbox}
\usepackage{comment}
\usepackage[accsupp]{axessibility}  % Improves PDF readability for those with disabilities.
\usepackage{fancyhdr}

\usepackage{tabularx}

\begin{document}

% \title{Accepted to 2023 CVPRW on Event-Based Vision https://tub-rip.github.io/eventvision2023/}

\title{\textcolor{red}{\fontsize{16}{18}\selectfont{Accepted to 2024 European Conference on Computer Vision} \newline }\newline \newline \fontsize{24}{18}\selectfont{Asynchronous Bioplausible Neuron for SNN for Event Vision}}

\author{Sanket Kachole$^{1}$ \hspace{0.5cm} Hussain Sajwani$^{2,3}$ \hspace{0.5cm} Fariborz Baghaei Naeini$^{1,4}$  \hspace{0.5cm} Dimitrios Makris$^{1}$   \hspace{0.5cm} Yahya Zweiri$^{2,3}$
\\
Department of Computer Science, Kingston University, London, UK$^1$  %\hspace{0.5cm} Ipsotek, an Eviden Company, London $^3$ \hspace{0.5cm} 
\\Advanced Research
and Innovation Center (ARIC), Khalifa University, Abu Dhabi, UAE$^2$ 
\\
Department of Aerospace Engineering, Khalifa University of Science and
Technology, Abu Dhabi, UAE$^3$
\\Ipsotek, an Eviden Company, London$^4$
\\
{\tt\small \{K1742163,f.baghaeinaeini,d.makris\}@kingston.ac.uk$^1$ \hspace{0.25cm} \{hussain.sajwani,yahya.zweiri\}@ku.ac.ae$^2$} }

% \fontsize{12}{18}\selectfont\markboth{{A\MakeLowercase{ccepted to} 2023 CVPRW \MakeLowercase{on} E\MakeLowercase{vent} B\MakeLowercase{ased} V\MakeLowercase{ision}}  \href{https://tub-rip.github.io/eventvision2023/}{\MakeLowercase{https://tub-rip.github.io/{eventvision2023}/}}}%

% \chead{\textbf{Accepted to 2023 CVPRW on Event-Based Vision https://tub-rip.github.io/eventvision2023/}}
%\chead{\href{https://github.com/sanket0707/GNN-Mixer.git}{https://github.com/sanket0707/GNN-Mixer.git}}

%Remember, if you use this you must call \IEEEpubidadjcol in the second

\maketitle

\begin{abstract}

Spiking Neural Networks (SNNs) offer a biologically inspired approach to computer vision that can lead to more efficient processing of visual data with reduced energy consumption. However, maintaining homeostasis within SNNs is challenging,  as it requires continuous adjustment of neural responses to preserve equilibrium and optimal processing efficiency amidst diverse and often unpredictable input signals. In response to these challenges, we propose the Asynchronous Bioplausible Neuron (ABN), a dynamic spike firing mechanism that offers a simple yet potent auto-adjustment to variations in input signals. Its parameters, Membrane Gradient (MG), Threshold Retrospective Gradient (TRG), and Spike Efficiency (SE), make it stand out for its easy implementation, significant effectiveness, and proven reduction in power consumption, a key innovation demonstrated in our experiments. Comprehensive evaluation across various datasets demonstrates ABN's enhanced performance in image classification and segmentation, maintenance of neural equilibrium, and energy efficiency. 
\end{abstract}

\section{Introduction}
\label{sec:intro}

% 1. General Context (importance of Robotic grasping, applications, need for segmentation),
Computer vision has experienced significant success over the last years due to the advancement of Artificial Neural Networks such as Convolutions Neural Networks (CNNs \cite{Kachole2020AHoneybees, sanket20163Tracker
} and Vision Transformers \cite{kachole2024bimodal, kachole2023bimodal, sanket20163Tracker}. Recently, the attraction of Spiking Neural Networks (SNNs) has been growing, due to their potential in real-time sensory processing, closer imitation of biological brain functionalities, and low power consumption. The advent of event-based cameras further accelerates this evolution, offering a promising path to harness the strengths of SNNs and address the challenges faced by traditional vision systems \cite{kachole2024asynchronous, naeini2022event}. This combination paves the way for more energy-efficient and adaptive solutions in computer vision  \cite{Pozo2010UnravelingPlasticity}.

\begin{figure*}[h!]
\centering
\begin{tabular}{|c|}
\hline
Overview of Standard Spiking Encoding and Membrane Threshold Methods  \\
\hline    
\subfloat{\includegraphics[width = 1\textwidth]{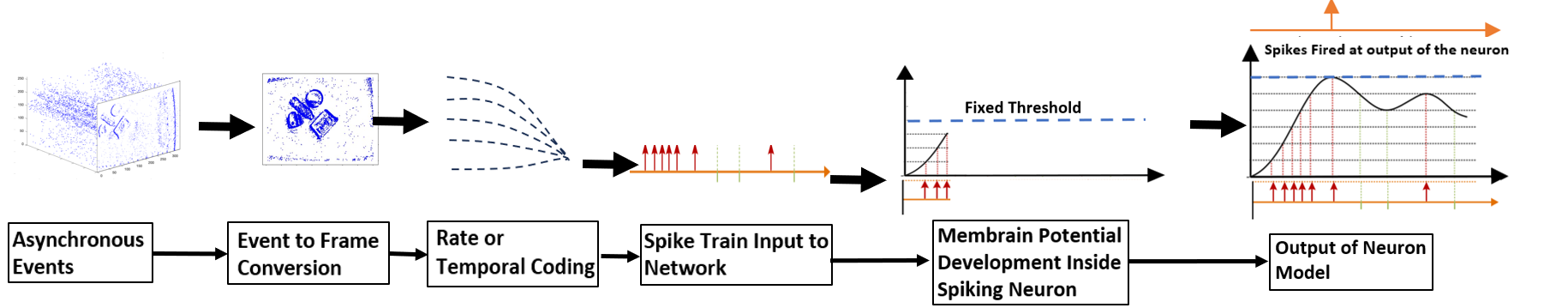}}  \\
\hline    
Our Encoding Free and Dynamic Membrane Threshold Method    \\
\hline
\subfloat{\includegraphics[width = 1\textwidth]{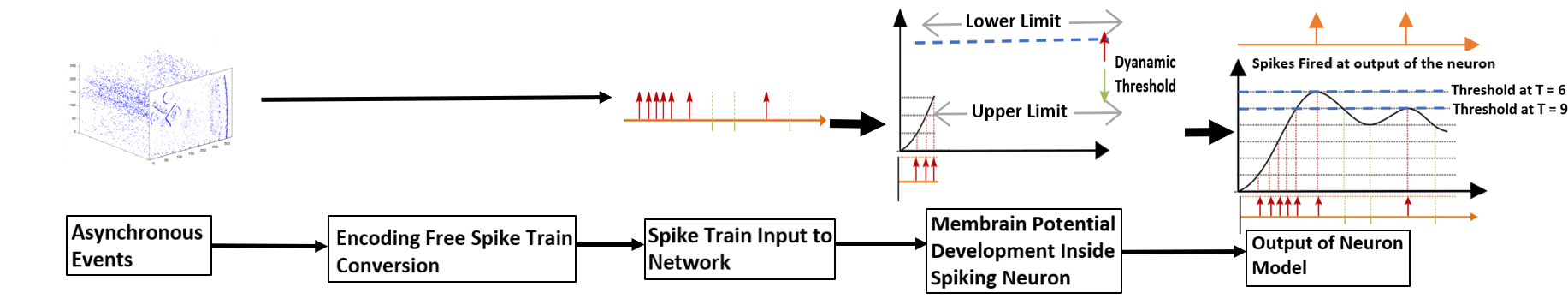}}  \\
\hline
\end{tabular}
\caption{Comparative Visualization of SNNs approaches: The top image presents the overview of the SNN methodology showcasing the conventional processes of asynchronous event capture, event-to-frame conversion, and fixed thresholding in neuronal spike response. In contrast, the bottom image illustrates an overview of the proposed method, highlighting the explicit encoding of events to spikes and implementing a dynamic thresholding mechanism for neuronal spiking. This juxtaposition underscores the novel methodology proposed.}
\label{tab:Introduction_Part_1}
\end{figure*}

SNNs' performance benefits from homeostasis, i.e. maintaining a stable neuron activity level, neither spiking too often nor staying too quiet \cite{Cazalets2023AnCombination, Yeung2004SynapticModel}. To achieve this, standard methods adjust parameters like fixed firing thresholds or synaptic weights. These strategies require fine-tuning, and may not be versatile across network architectures. The fixed thresholds in spiking neurons, as shown in Fig. \ref{tab:Introduction_Part_1}, can be sensitive to initial conditions, leading networks to be overly active or dormant. The inflexibility in adapting to varying inputs makes neurons either too sensitive or unresponsive, affecting negatively the performance. It also compromises energy efficiency, with either high power usage from excessive firing or missed data from sparse firing.

% 3. Proposed approach (multiple sensors), 

To address the above challenges, we introduce the Asynchronous Bioplausible Neuron (ABN) function, which can be incorporated into the two most prominent bioplausible neuron models: the Spike Response Model (SRM) \cite{Gerstner1995TimeModels.} and the Leaky Integrate-and-Fire (LIF) model \cite{Gerstner2002SpikingModels}. ABN is comprised of three novel elements  MG, TRG, and SE. In the MG component, we are the first to propose calculating gradients to quantify the movement of membrane potential, akin to the `rate of membrane potential change' observed in biological neurons, which reflects how neurons dynamically adjust their potential in response to inputs \cite{sanes1993activity}. This provides the direction of the movement necessary to maintain the neuron's output firing. In TRG, we are the first to implement inertia in threshold movement using a sliding window to control bursts of spikes, offering network-independent threshold control. This mirrors the `adaptive threshold' or `activity-dependent threshold modulation' observed in biological neurons, where firing thresholds adapt based on historical activity to maintain stability \cite{lu2003bdnf}. Although the principle of the SE element is adopted from existing literature, we are the first to implement it using a sliding window mechanism and asynchronous event data, relating it to `neuronal efficiency' or `firing efficiency,' which measures how effectively a neuron converts electrical inputs into action potentials, a crucial factor in preventing response saturation \cite{yu2017energy}.

We validate the effectiveness of our approach using two SNNs for object segmentation and image classification tasks under both normal and various degraded conditions, such as low light, high occlusion, etc. Further, our method is tested on six different datasets: N-MNIST \cite{Orchard2015ConvertingSaccades}, ESD-1, ESD-2 \cite{Huang2023AEnvironment , huang2024neuromorphic}, DVS128-Gesture \cite{Amir2017ASystem}, N-ImageNet \cite{kim2021n} and CIFAR-10 DVS \cite{Li2017CIFAR10-DVS:Classification} to demonstrate its general applicability. Additionally, we conducted an extensive study of the homeostasis (i.e. stability test of neuron firing), an ablation study to understand the impact of elements of the proposed function, and compared the power consumption with state-of-the-art methods.

\begin{figure}[t]
\centering
\fbox{
  \parbox{0.7\columnwidth}{
    \centering
    SOTA Thresholding Method \\
    \includegraphics[width=0.7\columnwidth]{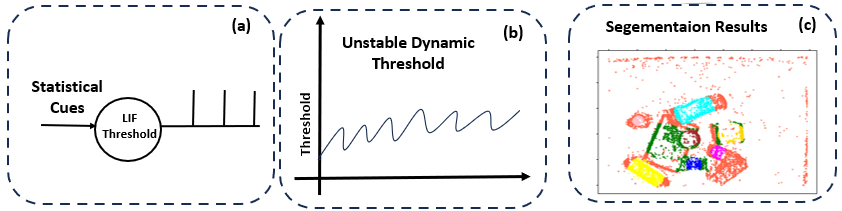}
  }
}
\\[1ex] % Adds extra space between the two boxes
\fbox{
  \parbox{0.7\columnwidth}{
    \centering
    Our Method \\
    \includegraphics[width=0.7\columnwidth]{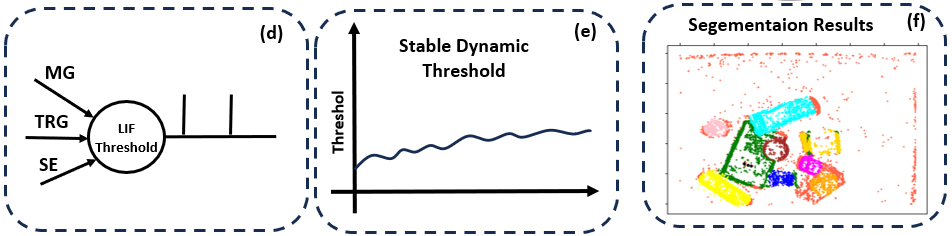}
  }
}
\caption{Dynamic thresholding for SNNs: Existing methods are based on the LIF neuron which uses statistical cues for thresholding (a), however they result in unstable thresholds (b) and consequently low accuracy for segmentation methods (c). Our proposed method incorporates weighted inputs in the form of MG, TRG, and SE (d), which results in a smoothed dynamic threshold (e), and improved segmentation (f).}
\label{fig:Introduction_DynamicThreshold}
\end{figure}

% 5. Paper outline.

The remainder of this paper is structured as follows: Section \ref{section : Literature} discusses related work, Section \ref{section : Methodology} details the proposed methodology, Section \ref{section : Results} presents experimental results, and Section \ref{section : Conclusion} concludes and discusses future research.

%%%%%%%%%%%%%%%%%%%%%%%%%%%%%%%%%%%%%%%%% Literature Review %%%%%%%%%%%%%%%%%%%%%%%%%%%%%%%%%%%%%%%%%%%

\section{Related Work}
\label{section : Literature}

\subsection{Computer vision applications of SNNs}
\label{subsection : Application of Spiking Neural Network}

The exploration of SNNs for object segmentation and image classification has ignited widespread interest in the field, as the SNNs replicate the biological neuronal system to provide efficient and effective solutions for this task. Initial works in object segmentation leaned on feed-forward SNN architectures, utilizing the spike-timing-dependent plasticity (STDP) learning rule for unsupervised learning of features \cite{Deng2021STDPClassification}. The Backpropagation Through Time (BPTT) method introduced in \cite{Kim2022BeyondSegmentation} for training SNNs, offers a more structured and supervised learning approach that significantly enhances feature extraction and classification accuracy, yet it is limited by its high computational complexity, substantial memory requirements, susceptibility to vanishing and exploding gradients, challenges in capturing long-term dependencies, and inefficiency in real-time learning applications. The introduction of Convolutional Spiking Neural Networks (CSNNs) marked a significant development in image classification. By incorporating convolutional layers into the network, these models could better handle the spatial complexities inherent in images, significantly enhancing segmentation performance. Concurrently, they maintained the temporal dynamics native to SNNs, enabling efficient processing of streaming data \cite{Li2023ANetworks}. Recently, there has been a surge in the application of recurrent spiking neural networks (RSNNs) to image classification. These networks leverage feedback connections, enhancing the network's ability to capture temporal dependencies in the data. Furthermore, their capacity to retain information over extended periods makes them particularly suited to segmentation tasks involving sequential or video data \cite{Bohnstingl2022Biologically-inspiredHardware}.

The memory-based methods such as those utilizing LSTM and RNN principles, highlighted by \cite{bellec2018long}
and \cite{bellec2020solution}, offer advancements yet face limitations due to their application across entire networks, constraining individual neuron-based threshold modeling. Similarly, Spiking Neural Networks (SNNs) have been applied to tasks like segmentation and object detection with promising results as demonstrated by \cite{zhang2023accurate} and \cite{zhang2023automotive}; however, these models often draw from Artificial Neural Network (ANN) principles, potentially veering away from the bio-plausibility sought to closely mimic biological brain functions. These developments underscore the challenges in achieving nuanced modeling and bio-inspired design within neural network research, highlighting the need for approaches that balance technical efficacy with the intricacies of biological mechanisms. Generative networks \cite{leng2018spiking} in event-based vision struggle due to the asynchronous and unpredictable nature of the data, leading to performance issues in scenarios with sporadic events. This challenge highlights the complexity of creating bio-plausible models that maintain stability and accuracy in the face of such data variability.

\subsection{Dynamic Thresholding in SNNs}
\label{subsection : Dynamic Thresholding in Spiking Neural Networks}

Research on SNNs has delved deeply into dynamic thresholding methods to bolster adaptability. For instance, while predictive coding emphasizes anticipating input patterns, it overlooks the spike timing significance, an aspect STDP-based adjustments address \cite{Rao1999PredictiveEffects}, \cite{Makhlooghpour2016HighModel}. The holistic approach of network activity regulation contrasts with the inherent refractory period modulation, focusing on preventing rapid neuron firing \cite{WangRegulationModel}. Furthermore, spiking models like SRM and LIF have advanced neuron dynamics representations \cite{Gerstner1995TimeModels., Gerstner2002SpikingPlasticity}. However, gaps remain, especially in dynamic threshold modulation. \cite{Hao2018ARuleb} introduced a method utilizing a scaling factor to adjust the threshold in response to increased input. Yet, since it was tailored specifically for incremental spike membrane potential, it faltered when faced with variations in membrane decays. On the other hand, \cite{Shaban2021AnImplementation.} adopted exponential functions to dictate decay. This approach, however, resulted in fluctuating thresholding levels, failing to ensure a smooth threshold curve. Meanwhile, \cite{Kim2021SpikingUpdate.} ventured into a firing counter-based threshold adjustment. While innovative, it necessitated meticulous fine-tuning and exhaustive experimentation, leaving the determination of an optimal count mired in ambiguity.

Previous research \cite{naeini2022event} may not addressed all the dynamics in the input signal, especially in highly complex event-based vision signals. In the event-based vision, SNNs face unique challenges: Neurons within the same layer exhibit varying event counts; some neurons experience more events than others. Consequently, applying uniformly the BDETT method \cite{Ding2022BiologicallyNetworks}, which utilizes average membrane potential of the neurons in a layer to predict the threshold of all the neurons in that layer, fails to reflect the individual dynamics of each neuron. 
This paper addresses notable research gaps. No prior studies have demonstrated that a bioinspired dynamic threshold can stabilize spike firing in semantic segmentation tasks. Additionally, these studies typically employ simplistic datasets such as NMNIST \cite{Orchard2015ConvertingSaccades} and rarely evaluate system robustness under challenging conditions like motion blur or low light, vital for assessing the system's homeostasis \cite{Aydin2023APerception}.

%%%%%%%%%%%%%%%%%%%%%%%%%%%%%%%%%%%%%%%%% Methodology %%%%%%%%%%%%%%%%%%%%%%%%%%%%%%%%%%%%%%%%%%%%%%

\section{Methodology}
\label{section : Methodology}

\subsection{Pre-requisite}
\label{subsection : Pre-requisite}

\subsubsection{Event-based cameras}
Event-based vision cameras are designed to register variations in logarithmic light intensities by recording individual pixel-level changes, referred to as events. These events formulate a continuous asynchronous event stream, which can be mathematically articulated as a series of ordered tuples. A tuple corresponds to an event $i$ and includes its spatial coordinates $(x_i, y_i)$ in the image frame, timestamps $t_i$, and polarity value $z_i$ \cite{Lichtsteiner2008ASensor}: 
\begin{equation}
    {(x_1, y_1, t_1, z_1), (x_2, y_2, t_2, z_2), ..., (x_n, y_n, t_n, z_n)}
\end{equation}

\subsubsection{Leaky integrate-and-fire}
The LIF model, as shown in Fig. \ref{tab:Introduction_Part_1}, is characterized by its linear integration of inputs until a threshold \( \Theta(t)\) is reached, causing its firing \cite{Gerstner2002SpikingModels}. Mathematically, the sub-threshold dynamics of the membrane potential \( V(t) \) of an LIF neuron can be described by the following differential equation:

\begin{equation}
\tau_m \frac{dV(t)}{dt} = - (V(t) - E_L) + R_m I(t)
\end{equation}

\noindent
where \( \tau_m \) is the membrane time constant that dictates the speed at which the neuron responds to inputs, \( E_L \) is the leak potential or resting potential representing the equilibrium potential of the neuron in the absence of any input, \( R_m \) is the membrane resistance, and \( I(t) \) represents the total synaptic input to the neuron that might come from other neurons or external sources. When the membrane potential \( V(t) \) reaches the threshold \( \Theta(t)\), the neuron emits a spike and then resets its potential to a reset value \( V_{\text{reset}} \), typically below \( \Theta(t)\). Additionally, after spiking, the neuron enters a refractory period for a duration \( T_{\text{ref}} \) during which it cannot fire again. The "leaky" aspect of the LIF model comes from the term \( (V(t) - E_L) \) in the differential equation. This term ensures that in the absence of any input, the neuron's potential will gradually revert to the leak potential \( E_L \).

\subsubsection{Spike response model}
In the SRM \cite{Gerstner1995TimeModels.}, incoming spike trains \( s_i(t) \) are initially transformed into a spike response signal by convolving them with a spike response kernel \( \varepsilon(\cdot) \), denoted as \( (\varepsilon \ast s_i)(t) \). Subsequently, this spike response signal is scaled by a synaptic weight \( w_i \). The depolarization process involves summing up these scaled spike response signals across all incoming spikes, mathematically represented as \( \Sigma_i w_i (\varepsilon \ast s_i)(t) \). The SRM also incorporates a mechanism for hyperpolarization when a spike \( s(t) \) is triggered, characterized by a refractory potential \( (\zeta \ast s)(t) \), where \( \zeta(\cdot) \) serves as the refractory kernel. This dual-component approach enables SRM to capture both excitatory and inhibitory dynamics in neuronal spike activity.

\begin{figure*}
    \centering
    \includegraphics[width=\textwidth, keepaspectratio]{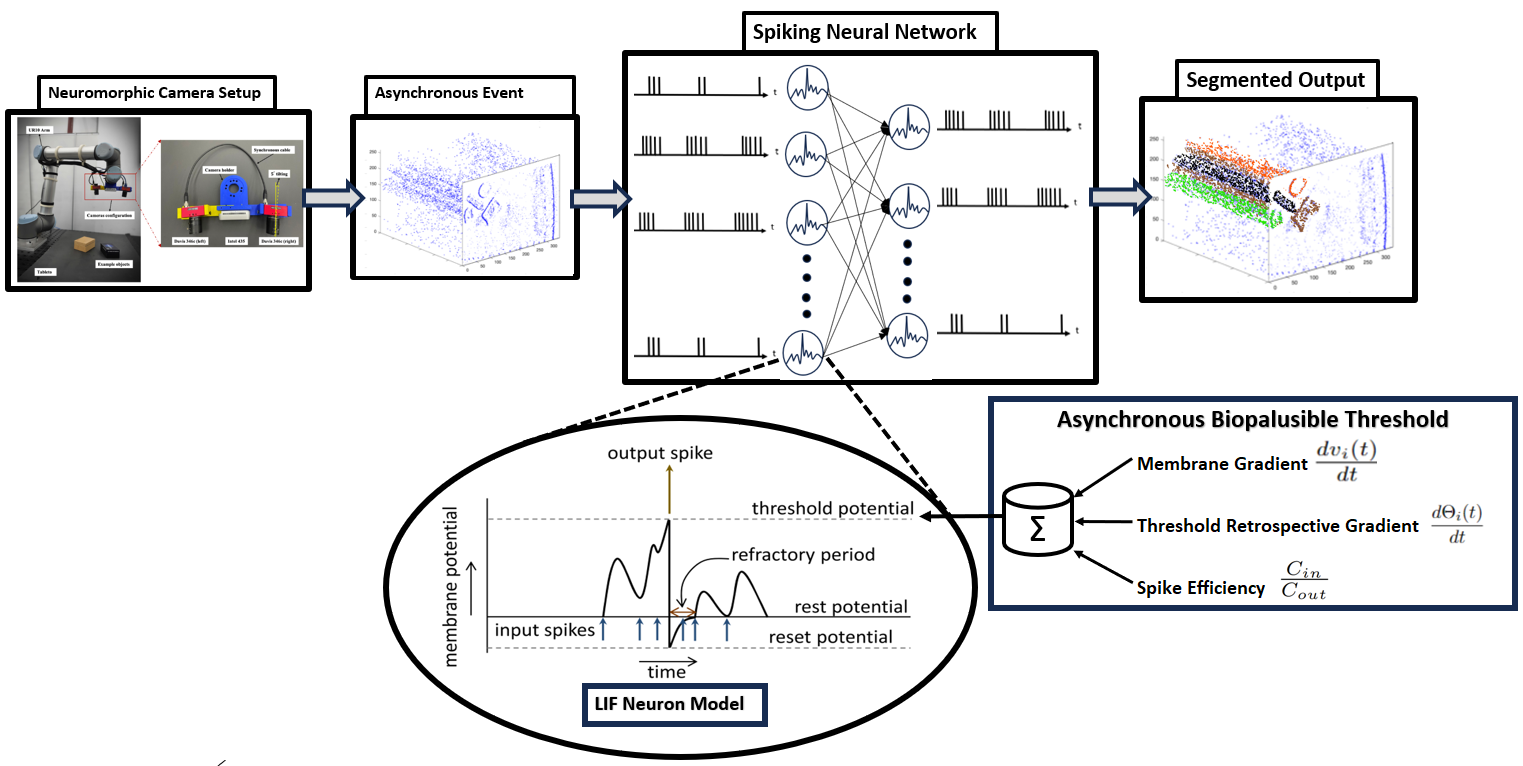}
    \caption{Schematic Diagram of ABN. The DVS camera outputs data in the form of asynchronous events. It is passed into a SNN with neurons in the first layer equal to the number of pixels on the camera. Each neuron adopts the proposed ABN where it takes the weighted input from the MG, TRG, and SE to control the dynamic threshold.}
    \label{fig:ABN}
\end{figure*}

\subsubsection{Asynchronous events to spike conversion}
\label{subsubsection : Asynchronous events to spike conversion}

Traditionally, rate coding and temporal coding \cite{Kim2022RateNetworks}, \cite{Lehmann2021SpikingBiCMOS} are the popular methods to convert frames, from a traditional CMOS, into spiking trains.
In this study, similar to \cite{Abreu2023FlowHardware}, each event is treated as a spike to avoid the conversation of the input signal into a spike train.
%In this study. Each pixel on the sensor corresponds to a neuron in the SNN, and that neuron fires a spike at the times when the pixel generates an event.
This approach essentially uses a form of temporal coding, because the information is conveyed through the timing of the spikes. We have chosen to drop the polarity aspect of the events due to its high sensitivity, which can introduce noise and complicate the signal processing. Mathematically, E(t) is the event stream, $e = (x, y, t)$ is an individual event with coordinates $x$, y and timestamp t, and $S_i(t)$ is the spike train for the i-th neuron:

\begin{equation}
S_i(t) = 
\begin{cases} 
 \delta(t - e_t) & \text{if } e_x = x_i \text{ and } e_y = y_i \\
0 & \text{otherwise}
\end{cases}
\end{equation}

\noindent
Here, $ \delta(t - e_t)$ is the Dirac delta function, which represents a "spike" at  time $t$.

\subsection{Asynchronous Bioplausible Neuron }
\label{subsection : Asynchronous Biopalusible Neuron}

The proposed ABN is a weighted combination of the novel Membrane Gradient (MG), Threshold Retrospective Gradient (TRG), and Spike Efficiency (SE) for threshold adjustment, uniquely tailored for each neuron. This individualization is crucial as each neuron receives a unique frequency of events, necessitating adaptation to input variations and ensuring the maintenance of firing stability, irrespective of the layer. Crucially, ABN's design embodies simplicity and functions as a lightweight adjustment, seamlessly integrating into existing frameworks without the need for complex modifications. The details of each component are discussed in the subsequent subsections below. 
The mathematical equation of the ABN is:

\begin{equation}
\begin{aligned}
\Theta_{i}(t+1) = \Theta_{i}(t) + k_1 MG_{i}  - k_2 TRG_{i} + k_3 SE_{i}
\end{aligned}
\end{equation}

The equation models the dynamic adjustment of the neuron's firing threshold at the next time step \( t+1 \) based on various factors. In this equation, \( \Theta(t) \) is the existing threshold at time \( t \), and \( k_1, k_2, k_3 \) are weights that determine the relative influence of each term. The term \( k_1 \cdot MG_{i} \) increases the threshold when the membrane potential is rapidly changing, preventing the neuron from firing too easily. Conversely, the term \( -k_2 \cdot TRG_{i} \) decreases the threshold if there has been a historical trend of rapid threshold increases. Lastly, the term \( k_3 \cdot SE_{i} \) takes into account the neuron's recent spiking activity. Specifically, if the neuron is overly active, this term will make it more difficult for the neuron to fire in the future, thereby ensuring efficiency. Overall, this equation encapsulates a multi-faceted approach to dynamically adapting the neuron's firing threshold.

\subsubsection{Membrane Gradient (MG)}
\label{subsubsection : Membrane Gradient}

The rate of change of the membrane potential is a critical parameter for predicting a dynamic threshold in SNNs. It serves as an indicator of the neuron's responsiveness to incoming spikes and plays an essential role in governing the neuron's firing behavior. Positive correlations between dynamic thresholds and membrane potentials have been observed in several areas of diverse biological nervous systems. In \cite{Fontaine2014Spike-thresholdVivo.}, the average membrane potential is correlated with an average threshold to predict the dynamic threshold. Instead, in this work we correlate the rate of change of membrane potential with a threshold. A high rate of change often signifies that the neuron is receiving a burst of spikes, necessitating a higher threshold to maintain neuronal homeostasis. Conversely, a lower rate of change could indicate that the neuron is less active, allowing for a lower dynamic threshold to be sensitive to sparse inputs. Incorporating the rate of change of the membrane potential into the calculation of a dynamic threshold ensures that the SNNs can adapt to varying synaptic activities, thereby emulating the adaptability and energy efficiency observed in biological neurons.

The BDETT method presented in \cite{Ding2022BiologicallyNetworks} calculates the dynamic threshold using the average of membrane potentials and thresholds. In a layer with varied neuronal activity, not all neurons are always active. Including inactive neurons in this calculation can pull the average value lower. Applying this reduced average threshold to highly active neurons may result in excessive spiking, leading to overly aggressive neuronal activity. In contrast, in this work, we adopt the neuron-specific derivative term, denoted as $\eta \cdot \frac{dv_{i}(t)}{dt}$, which captures the rate of change of membrane potential of neuron $i$ at time $t$, denoted by $v_{i}(t)$. To numerically compute the derivative, a finite difference approximation is employed. This involves the subtraction of the membrane potential at the previous timestep from the current potential, followed by division by the timestep size $\Delta t$. The scaling factor $\eta$, which is set to a value less than 1, reduces the rate of change to ensure that the rate of change of the threshold remains less than the rate of change of the potential. The explicit equation for the derivative term is given as:

\begin{equation}
MG_{i} = \frac{d v_{i}(t)}{dt} \approx \frac{v_{i}(t) - v_{i}(t - 1)}{\Delta t}
\end{equation}

\subsubsection{Threshold Retrospective Gradient (TRG)}
\label{subsubsection : Threshold Retrospective Gradient}

Biological systems exhibit a negative correlation with the preceding rate of depolarization, which indicates the excitatory status. We propose considering a historical rate of change for the threshold \( \Theta(t)\). This is because focusing solely on the preceding rate of depolarization as implemented in \cite{Ding2022BiologicallyNetworks} is insufficient and can result in an unstable or overly sensitive dynamic threshold. This approach lacks memory, reacting too quickly to transient changes without considering longer-term neuronal activity patterns. Consequently, this can result in excessive threshold fluctuations and potentially amplify noise, undermining network stability and reliability. Therefore, we take into account the average of the historical rate of change of the dynamic threshold. Utilizing the historical rate of change of the dynamic threshold to calculate the current threshold can provide the neuron with inertia or memory of its past states. This can be useful in situations where the neuron should react not only based on the current state but also considering its past behavior. In discrete terms, if \( \Theta(t)\) is the threshold at time \( t \), then the rate of change at \( t \) is:

\begin{equation}
G_{i}(t) = \frac{\Theta_{i}(t) - \Theta_{i}(t-1)}{\Delta t}
\end{equation}

The historical rate of change is an average of these rates over a recent window of time, \( N \) time steps:

\begin{equation}
\overline{G}_{i}(t) = \frac{1}{N} \sum_{j=0}^{N-1} G_{i}(t-j)
\end{equation}

In order to ensure that the older rates influence the current threshold less within  the sliding window, a decay factor $\alpha, ( 0 < \alpha < 1 )$, is used:

\begin{equation}
TRG_{i} = \frac{1}{N} \sum_{j=0}^{N-1} \alpha^j G_{i}(t-j)
\end{equation}

\noindent
with values closer to 0 suppressing more the older rates.

In summary, leveraging the historical rate of change of the dynamic threshold has introduced a form of temporal inertia to the neuron, creating a relationship to patterns in input data over time. Importantly, it is especially useful in scenarios where temporal patterns are crucial.

\subsubsection{Spike Efficiency (SE)}
\label{subsubsection : Spike Efficiency}

The ratio of spike output \( C_{\text{out}} \) to spike input \( C_{\text{in}} \) can serve as a crucial metric for determining the dynamic threshold in SNNs. Intuitively, this ratio indicates how efficiently a neuron converts incoming spikes into outgoing spikes. A high SE suggests that the neuron is highly responsive to incoming activity, and could therefore benefit from a high dynamic threshold to maintain energy efficiency and reduce the likelihood of over-saturation. Conversely, a low SE implies that the neuron is less responsive, possibly requiring to lower a dynamic threshold to avoid unnecessary firing and thereby preserve neuronal and computational resources. This ratio is mathematically represented as:

\begin{equation}
r_{i}(t) = \frac{C_{\text{out}, i}(t)}{C_{\text{in}, i}(t)}
\end{equation}

When using a sliding window method, we only consider \( N \) previous time steps to calculate \( C_{\text{in}} \) and \( C_{\text{out}} \). This allows the threshold to adapt quickly to recent changes in spike rates.

\begin{equation}
SE_{i} = \bar{r}_{i}(t) = \frac{\sum_{j=t-N}^{t} C_{\text{out}, i}(j)}{\sum_{j=t-N}^{t} C_{\text{in}, i}(j)}
\end{equation}

Incorporating SE into the dynamic threshold calculation enables the SNN to adjust its sensitivity according to the context of incoming activity. This inclusion ensures a more nuanced and adaptive network behaviuor, aligning closely with biological plausibility while maintaining computational efficiency. It allows the network to operate in a balanced regime, irrespective of the incoming spike rates, ensuring robust and stable performance across different input conditions. This adaptive mechanism is invaluable, especially when individual neurons within a network necessitate autonomous regulation of their activity, contingent on their functional significance or the unique input patterns they discern.

%%%%%%%%%%%%%%%%%%%%%%%%%%%%%%% Results %%%%%%%%%%%%%%%%%%%%%%%%%%%%%%%%%%%%%%%%%

\section{Results}
\label{section : Results}

\subsection{Experimental Protocol}
\label{subsection : Experimental Protocol}

We assess the effectiveness of the proposed ABN in two Computer Vision tasks: Image Classification for the N-MNIST \cite{Orchard2015ConvertingSaccades}, CIFAR10-DVS \cite{Li2017CIFAR10-DVS:Classification}, N-ImageNet \cite{kim2021n}, DVS128 gesture recognition \cite{Amir2017ASystem} datasets, and Object Semantic Segmentation for the ESD-1 \cite{Huang2023AEnvironment} and ESD-2 \cite{Huang2023AEnvironment} datasets. The details of the datasets are explained in supplementary section B. The Spiking MLP employs Spatio-Temporal Backpropagation \cite{Ding2022BiologicallyNetworks} for training, allowing the network to consider both spatial and temporal information during the learning process. Although the input to the first layer should be a spike train, we can directly use asynchronous event-based vision data without requiring specific conversion. The initialization of network parameters, such as weights and thresholds, is crucial for stabilizing the network's firing activities. We adopt the initialization strategy as discussed in \cite{Wu2018Spatio-TemporalNetworks}. To make a fair comparison with  \cite{Ding2022BiologicallyNetworks}, we have used a similar experimental configuration. The model was trained on a GPU PC featuring 128GB of memory, an Intel Xeon W-2155 Processor, and NVIDIA Quadro RTX 8000 48GB graphics cards, housed in a Lenovo ThinkStation P520. While this is essential for assessing the efficacy of the proposed method, implementing it directly on neuromorphic hardware may require additional engineering efforts, which are beyond the scope of this work. Custom code was written to develop the LIF and SRM within this framework. Synaptic weights are initialized randomly, and a decay constant, \(\tau_1\), is set to 0.1 ms. The weights for constants \(k_1\), \(k_2\), and \(k_3\) are experimentally selected as 0.25, 0.50, and 0.25, respectively (see supplementary document).

\subsection{Comparative Results across Diverse Datasets}
\label{subsection : Generalization Evaluation}

\begin{table}[t]
    \caption{Classification Accuracy across Various Datasets and Models}
    \begin{adjustbox}{width=0.42\textwidth}
    \label{tab:classification}
    \centering
    \begin{tabularx}{\columnwidth}{l|cc|cc|cc}  
        \toprule
        Method & \multicolumn{2}{c|}{N-MNIST} & \multicolumn{2}{c|}{DVS128} & \multicolumn{2}{c}{CIFAR10} \\
        \midrule
        & SRM & LIF & SRM & LIF & SRM & LIF \\
        \hline
        Spiking RBM \cite{Neftci2014Event-drivenSystems} & 92.1 & 93.16 & 87.38 & 90.21 & 83.04 & 86.25 \\
        Spiking MLP (BP)  \cite{OConnor2016DeepNetworks} & 94.52 & 97.66 & 90.72 & 93.4 & 86.23 & 90.06 \\
        Spiking MLP (STDP)  \cite{Ferre2018UnsupervisedSTDP} & 93.47 & 95 & 90.8 & 92.01 & 87.5 & 91.7 \\
        Spiking MLP (STBP) \cite{Wu2018Spatio-TemporalNetworks} & 97.13 & 98.89 & 92.54 & 93.64 & 86.23 & 91.73 \\
        DT1 \cite{Hao2018ARuleb} & 99.05 & 99.4 & 95.01 & 96.88 & 89.17 & 92.65 \\
        DT2 \cite{Kim2021SpikingUpdate.} & 98.13 & 98.24 & 92.54 & 95.54 & 89.38 & 91.47 \\
        BDETT \cite{Ding2022BiologicallyNetworks} & 99.15 & 99.45 & 94.09 & 96.05 & 91.61 & 93.5 \\
        ABN (Ours) & \textbf{99.23} & \textbf{99.48} & \textbf{95.64} & \textbf{98.74} & \textbf{93.5} & \textbf{94.74} \\
        \bottomrule
    \end{tabularx}
     \end{adjustbox}
    \label{tab: Classification Accuracy across Various Datasets and Models}
\end{table}

\begin{table}[t]
    \caption{Segmentation Accuracy across Various Datasets and Models}
    \begin{adjustbox}{width=0.48\textwidth}
    \label{tab:segmentation}
    \centering
    \begin{tabularx}{\columnwidth}{l|cc|cc}  
        \toprule
        Method & \multicolumn{2}{c|}{ESD-1} & \multicolumn{2}{c}{ESD-2} \\
        \midrule
        & SRM & LIF & SRM & LIF \\
        \hline
        Spiking RBM \cite{Neftci2014Event-drivenSystems} & 48.95 & 51.05 & 45.2 & 50.31 \\
        Spiking MLP (BP) \cite{OConnor2016DeepNetworks} & 49.53 & 54 & 45.5 & 52.42 \\
        Spiking MLP (STDP) \cite{Ferre2018UnsupervisedSTDP} & 52.63 & 58.84 & 46.7 & 49.44 \\
        Spiking MLP (STBP) \cite{Wu2018Spatio-TemporalNetworks} & 55.82 & 63.09 & 49.07 & 51.37 \\
        DT1 \cite{Hao2018ARuleb} & 57.55 & 61.45 & 50.13 & 54.59 \\
        DT2 \cite{Kim2021SpikingUpdate.} & 58.28 & 64.3 & 53.49 & 56.5 \\
        BDETT \cite{Ding2022BiologicallyNetworks} & 61.02 & 65.39 & 51.36 & 55.46 \\
        ABN (Ours) & \textbf{61.56} & \textbf{67.29} & \textbf{56.23} & \textbf{58.04} \\
        \bottomrule
    \end{tabularx}
    \end{adjustbox}
    \label{tab: Segmentation Accuracy across Various Datasets and Models}
\end{table}

The evaluation of our ABN method's generalization capabilities, across diverse datasets and neuron models is shown in Table \ref{tab: Classification Accuracy across Various Datasets and Models} and \ref{tab: Segmentation Accuracy across Various Datasets and Models}. Performance on the classification datasets was evaluated using classification accuracy and on the object segmentation using event accuracy. The metrics are detailed in the supplementary document section C. 
On the N-MNIST dataset, ABN achieved the highest classification accuracy (\textbf{99.23\%} with SRM and \textbf{99.48\%} with LIF), surpassing other methods like BDETT and DT1. In the ESD-1 and ESD-2 datasets, designed for robotic grasping tasks, ABN again outperformed others, showing particularly strong performance in the ESD-2 dataset with unknown objects (\textbf{56.23\%} with SRM and \textbf{58.04\%} with LIF). For the DVS128 Gesture, CIFAR10-DVS and N-ImageNet datasets, ABN recorded the highest scores \textbf{95.64\%}, \textbf{93.5\%} and \textbf{56.73\%} with SRM and in LIF \textbf{98.74\%}, \textbf{94.74\%} and \textbf{57.04\%} , indicating its robustness and effectiveness. This comprehensive performance underscores the superior adaptability and effectiveness of the ABN method across a variety of datasets and computer vision tasks.

\subsection{Evaluation on Degraded Input on ESD-1}
\label{subsection : Evaluation on Degraded Input with Known Objects ESD- 1}

\begin{table}[t]
    \centering
    \caption{Segmentation accuracy of known objects in various conditions.}

   \begin{adjustbox}{width=0.48\textwidth}

    \centering
      {\fontsize{10}{12}\selectfont
    % \begin{tabular}{|c|c|c|c|c|c|c|}      
    \begin{tabular}{lccccc}
  
  \specialrule{.15em}{.1em}{.1em}  
        \multicolumn{6}{c}{Exp 1: \textbf{varying clutter objects}, Bright light, 62cm height,  Rotational motion, 0.15 m/s speed } \\
  \specialrule{.1em}{.1em}{.1em} 
        Method & 2 Obj & 4 Obj & 6 Obj & 8 Obj & 10 Obj   \\  %& \begin{tabular}{c} Overlapped \\ Events \\ (\% of FP)\end{tabular}
  \specialrule{.1em}{.1em}{.1em}

        Spiking MLP (STBP) \cite{Wu2018Spatio-TemporalNetworks} & 74.16\% & 73.05\% & 66.79\% & 66.42\% & 54.42\% \\
        DT1 \cite{Hao2018ARuleb} & 77.56\% & 74.65\% & 68.20\% & 68.70\% & 55.49\% \\
        DT2 \cite{Kim2021SpikingUpdate.} & 78.46\% & 75.42\% & 69.28\% & 69.62\% & 56.16\% \\
       BDETT \cite{Ding2022BiologicallyNetworks} & 79.07\% & 78.03\% & 74.14\% & 69.83\% & 58.28\% \\
        ABN (ours) & \textbf{81.49\%} & \textbf{79.74\%} & \textbf{74.97\%} & \textbf{72.16\%} & \textbf{60.03\%} \\
        
   \specialrule{.15em}{.1em}{.1em}

   \specialrule{.15em}{.1em}{.1em}  
        \multicolumn{6}{c}{Exp 2: 6 Objects, \textbf{varying lighting conditions}, 62cm height, Rotational Motion, 0.15 m/s speed.} \\
  \specialrule{.1em}{.1em}{.1em} 
        Method & Bright Light & Low light &  &  &    \\  %& \begin{tabular}{c} Overlapped \\ Events \\ (\% of FP)\end{tabular}

        Spiking MLP (STBP) \cite{Wu2018Spatio-TemporalNetworks} & 58.73\% & 61.59\% \\
        DT1 \cite{Hao2018ARuleb} & 61.32\% & 62.91\% \\
        DT2 \cite{Kim2021SpikingUpdate.} & 61.24\% & 64.53\% \\
        BDETT \cite{Ding2022BiologicallyNetworks} & 63.92\% & 66.03\% \\
        ABN (Ours) & \textbf{65.98\%} & \textbf{67.42\%} \\

   \specialrule{.15em}{.1em}{.1em}

   \specialrule{.15em}{.1em}{.1em}  
        \multicolumn{6}{c}{Exp 3: 6 Objects, Bright Light, 62cm height, \textbf{Varying directions of motion}, 0.15 m/s speed.} \\
  \specialrule{.1em}{.1em}{.1em} 
        Method & Linear & Rotational & Partial Rotational &  &    \\  %& \begin{tabular}{c} Overlapped \\ Events \\ (\% of FP)\end{tabular}

         Spiking MLP (STBP) \cite{Wu2018Spatio-TemporalNetworks} & 48.09\% & 63.99\% & 67.43\% \\
        DT1 \cite{Hao2018ARuleb} & 49.09\% & 65.14\% & 69.31\% \\
        DT2 \cite{Kim2021SpikingUpdate.} & 50.43\% & 66.89\% & 69.79\% \\
        BDETT \cite{Ding2022BiologicallyNetworks} & 52.09\% & 68.60\% & 72.04\% \\
        ABN (Ours) & \textbf{55.50\%} & \textbf{69.45\%} & \textbf{73.54\%} \\

   \specialrule{.15em}{.1em}{.1em}

   \specialrule{.15em}{.1em}{.1em}  
        \multicolumn{6}{c}{Exp 4: 6 Objects, Bright Light, 62cm height, Rotational motion, \textbf{Varying speed}.} \\
  \specialrule{.1em}{.1em}{.1em} 
        Method & 0.15 m/s & 0.3 m/s &   0.1 m/s &  &    \\  %& \begin{tabular}{c} Overlapped \\ Events \\ (\% of FP)\end{tabular}
  \specialrule{.1em}{.1em}{.1em}   
  
        Spiking MLP (STBP) \cite{Wu2018Spatio-TemporalNetworks} & 53.16\% & 68.12\% & 56.29\% \\
        DT1 \cite{Hao2018ARuleb} & 54.56\% & 70.27\% & 60.16\% \\
        DT2 \cite{Kim2021SpikingUpdate.} & 55.27\% & 72.31\% & 59.63\% \\
        BDETT \cite{Ding2022BiologicallyNetworks} & 56.67\% & 75.09\% & 63.16\% \\
        ABN (Ours) & \textbf{59.63\%} & \textbf{76.72\%} & \textbf{63.56\%} \\

   \specialrule{.15em}{.1em}{.1em}

   \specialrule{.15em}{.1em}{.1em}  
        \multicolumn{6}{c}{Exp 5: 6 Objects, Bright Light, \textbf{Varying camera height}, Rotational motion, Varying speed.} \\
  \specialrule{.1em}{.1em}{.1em} 
        Method & 62 cm & 82 cm &    &   &    \\  %& \begin{tabular}{c} Overlapped \\ Events \\ (\% of FP)\end{tabular}
  \specialrule{.1em}{.1em}{.1em}   
  
        Spiking MLP (STBP) \cite{Wu2018Spatio-TemporalNetworks} & 62.93\% & 55.05\% \\
    DT1 \cite{Hao2018ARuleb} & 63.82\% & 59.08\% \\
    DT2 \cite{Kim2021SpikingUpdate.} & 65.63\% & 59.93\% \\
    BDETT \cite{Ding2022BiologicallyNetworks} & 65.82\% & 59.00\% \\
    ABN (ours) & \textbf{68.40\%} & \textbf{63.27\%} \\

   \specialrule{.15em}{.1em}{.1em} 
   \specialrule{.15em}{.1em}{.1em}

    \end{tabular}}
    \end{adjustbox}
    \label{tab: Segmentation Accuracy in Various conditions}
     \setlength{\belowcaptionskip}{-12pt} 
\end{table}

We conducted an experiment focused on evaluating the segmentation performance of various state-of-the-art methods, and compared them to our proposed ABN, under conditions of degraded inputs. This experiment was essential to mirror real-world scenarios affected by factors like low light, occlusion, variations in speed and distance from the object, and the direction of camera motion. These factors significantly impact the quality of input, making it crucial to determine the robustness of the proposed method in less-than-ideal conditions. Utilizing the ESD-1 dataset, we measured the event accuracy to gauge the performance. The results across various conditions, as depicted in Table \ref{tab: Segmentation Accuracy in Various conditions}, consistently show that the ABN method outperforms all other SNN methods in all tested scenarios. Clearly, the superior ability of ABN to maintain high segmentation accuracy despite the challenging input conditions, confirms its efficacy and potential for real-world applications.

\subsection{Homeostasis Evaluation}
\label{subsection : Homeostasis Evaluation}

Homeostasis in SNNs is crucial as it reflects the network's ability to maintain stable internal conditions despite external environmental variations. It is an important indicator of the network's reliability and efficiency, especially when dynamic input is considered.

\begin{table}[h]
    \centering
    \caption{Homeostasis Performance on ESD-1 Known Object }   
   \begin{adjustbox}{width=0.48\textwidth}
    \centering
      {\fontsize{10}{14}\selectfont
      \begin{tabular}{c|c|c|c|c|c|c|c}
       \specialrule{.15em}{.1em}{.1em}  
         & & $FR\_m$ & $\Delta FR\_m$ & $FR\_m\_std$ & $\Delta FR\_m\_std$ &  $FR\_s\_std$ & $\Delta FR\_s\_std$ \\

        \hline      
        \multirow{5}{*}{Ideal Condition}        
        & Spiking MLP (STBP) \cite{Wu2018Spatio-TemporalNetworks} & 0.631 & & 0.485 & & 0.001564 & \\
        & DT1 \cite{Hao2018ARuleb} & 0.524 & & 0.459 & & 0.001146 &  \\
        & DT2 \cite{Kim2021SpikingUpdate.} & 0.507 & &0.416 & & 0.000983 &  \\
        & BDETT \cite{Ding2022BiologicallyNetworks} & 0.436 && 0.403 & &0.001224 &  \\
        & ABN (Ours) & \textbf{0.385}  & & \textbf{0.253}  & & \textbf{0.000891}  &  \\
        
        \hline
        
        \multirow{5}{*}{Occlusion} 
        & Spiking MLP (STBP) \cite{Wu2018Spatio-TemporalNetworks} & 0.783 & 0.152 & 0.582 & 0.097 & 0.001396 & 0.000168 \\
        & DT1 \cite{Hao2018ARuleb} & 0.692 & 0.168 & 0.561 & 0.102 & 0.003963 & 0.002817 \\
        & DT2 \cite{Kim2021SpikingUpdate.} & 0.586 & 0.079 & 0.484 & 0.068 & 0.002724 & 0.001741 \\
        & BDETT \cite{Ding2022BiologicallyNetworks} & 0.47 & 0.034 & 0.357 & 0.046 & 0.002291 & 0.001067 \\
        & ABN (Ours) & \textbf{0.373}  & \textbf{0.012}  & \textbf{0.291}  & \textbf{0.038}  & \textbf{0.000784}  & \textbf{0.000107}  \\

        \hline

        \multirow{5}{*}{ Light}
        & Spiking MLP (STBP)\cite{Wu2018Spatio-TemporalNetworks} &0.786&0.155&0.594&0.109&0.003942&0.002378 \\      
         & DT1 \cite{Hao2018ARuleb} &0.698&0.174&0.405&0.054&0.002508&0.001362  \\
        & DT2 \cite{Kim2021SpikingUpdate.} &0.594&0.087&0.489&0.073&0.003641&0.002658  \\
        & BDETT \cite{Ding2022BiologicallyNetworks} &0.498&0.062&0.467&0.064&0.001157&0.000067  \\
        & ABN (Ours) & \textbf{0.405} & \textbf{0.02} & \textbf{0.295} & \textbf{0.042} & \textbf{0.000889}  & \textbf{0.000002}   \\

        \hline
        \multirow{5}{*}{Speed} 
        & Spiking MLP (STBP) \cite{Wu2018Spatio-TemporalNetworks} &0.746&0.115&0.582&0.097&0.001666&0.000102 \\      
 & DT1 \cite{Hao2018ARuleb} &0.62&0.096&0.406&0.053&0.001167&0.000021  \\
 & DT2 \cite{Kim2021SpikingUpdate.} &0.609&0.102&0.351&0.065&0.002839&0.001856  \\
 & BDETT \cite{Ding2022BiologicallyNetworks} &0.517&0.081&0.352&0.051&0.002077&0.000853  \\
 & ABN (Ours) & \textbf{0.305} & \textbf{0.08} & \textbf{0.206} & \textbf{0.047} & \textbf{0.000881} & \textbf{0.00001}   \\
 
        \hline
        \multirow{5}{*}{Motion Direction} 
        & Spiking MLP (STBP) \cite{Wu2018Spatio-TemporalNetworks}  &0.701&0.07&0.505&0.02&0.003798&0.002234 \\      
 & DT1 \cite{Hao2018ARuleb} &0.601&0.077&0.409&0.05&0.003461&0.002315  \\
 & DT2 \cite{Kim2021SpikingUpdate.} &0.415&0.092&0.397&0.019&0.002827&0.001844  \\
 & BDETT\cite{Ding2022BiologicallyNetworks}  &0.395&0.041&0.351&0.052&0.00133&0.000106  \\
 & ABN (Ours) & \textbf{0.378} & \textbf{0.007} & \textbf{0.243} & \textbf{0.01} & \textbf{0.000875} & \textbf{0.000016}   \\

        \hline
        \multirow{5}{*}{Size Variance} 
        & Spiking MLP (STBP) \cite{Wu2018Spatio-TemporalNetworks}  &0.759&0.128&0.533&0.048&0.003901&0.002337 \\       
& DT1 \cite{Hao2018ARuleb} &0.675&0.151&0.449&0.01&0.002263&0.001117  \\
 & DT2 \cite{Kim2021SpikingUpdate.} &0.551&0.044&0.45&0.034&0.001346&0.000363  \\
 & BDETT \cite{Ding2022BiologicallyNetworks} &0.481&0.045&0.354&0.049&0.002763&0.001539  \\
 & ABN (Ours) & \textbf{0.373} & \textbf{0.012} & \textbf{0.255} & \textbf{0.002} & \textbf{0.000879} & \textbf{0.000012}   \\
 
        \hline

 \end{tabular}
 } % <-- Close the brace here
    \end{adjustbox}
    \label{tab:homeostasis-performance}
    \setlength{\belowcaptionskip}{-12pt} 
\end{table}

In an experimental setup across all variations of the ESD-1 dataset, various statistical indicators were used to quantify the homeostasis of the host SNNs. Specifically, for a fair comparison with \cite{Ding2022BiologicallyNetworks} we employed \(FR_m\), \(FR_{m\_std}\), and \(FR_{s\_std}\) metrics, representing the average neuron firing rate over all trials, the mean of standard deviations of neuron firing rates over individual trials, and the variability of these standard deviations, respectively. Our proposed ABN model demonstrates significantly lower values across all three metrics and all conditions. These results suggest a stable and well-regulated network, highlighting the ABN model's capacity for homeostasis.

\subsection{Energy Consumption}
\label{subsection : Power Consumption}

Table \ref{tab:power-consumption} presents an analysis of power consumption, comparing CNN, SotA SNNs and the ABN using the DVS128 gesture recognition dataset. As proposed in \cite{Aydin2023APerception}, MAC (Multiply-Accumulate operations) and AC (Accumulate Count) were used to estimate power consumption. A detailed description of power consumption calculations is provided in the supplementary document. 

CNN models such as  \cite{Calabrese2019DHP19:Dataset} and \cite{Baldwin2023Time-OrderedCameras}, show a higher power consumption of 0.541W and 1.56W, respectively, despite their zero AC. In contrast, using SNN models leads to significantly lower power usage, with hybrid and SNN models \cite{Aydin2023APerception} consuming 0.404W and 0.053W, respectively. BDETT's SNN model \cite{Aydin2023APerception} further reduces this consumption to 0.046W. Notably, our proposed SNN model outperforms these by achieving the lowest power consumption of just 0.038W, even with a higher AC of 105, underscoring the efficiency of our model in terms of energy usage compared to both CNNs and other SNNs.

\begin{table}[h]
    \centering
    \caption{Power Consumption Comparison}
   \begin{adjustbox}{width=0.48\textwidth}
    \centering
      {\fontsize{4}{6}\selectfont    
    \label{tab:power-consumption}
    \begin{tabular}{|c|c|c|c|c|}
        \hline
        Method & Model & MAC & AC & Power (W) \\
        \hline
        Calabrese \cite{Calabrese2019DHP19:Dataset} & CNN & 285 & \textbf{0} & 0.541 \\
        Baldwin \cite{Baldwin2023Time-OrderedCameras} & CNN & 963 & \textbf{0} & 1.56 \\
        Asude \cite{Aydin2023APerception} & Hybrid & 235 & 81 & 0.404 \\

        Asude \cite{Aydin2023APerception} & SNN & 0.6 & 123 & 0.053 \\

        BDETT \cite{Ding2022BiologicallyNetworks} & SNN & 0.7 & 115 & 0.046 \\
        
        Ours & SNN & \textbf{0.4} & 105 & \textbf{0.038} \\
        \hline
    \end{tabular}
     } % <-- Close the brace here
    \end{adjustbox}
\end{table}

\section{Conclusion}
\label{section : Conclusion}

This paper introduces the Asynchronous Bioplausible Neuron (ABN) within a spiking multi-layer perceptron (MLP). This innovative spiking neuron incorporates a dynamic threshold based on Membrane Gradient (MG) for Spike Frequency Adaptation, Threshold Retrospective Gradient (TRG) for Burst Suppression, Spike Efficiency (SE) for Homeostasis. The spiking MLP, which utilizes STDP, serves as the architecture for implementing the proposed neuron. 

A methodology for object segmentation and image classification that employs a dynamic vision sensor and the ABN is also proposed. Remarkably, this system can directly process asynchronous, event-based vision signals without requiring any preprocessing. Its performance was rigorously evaluated on both conventional datasets (N-MNIST, DVS128 Gesture, N-ImageNet and CIFAR10-DVS) and specialized robotic grasping datasets (ESD-1, ESD-2). The method achieved state-of-the-art performance in terms of classification accuracy and event-wise accuracy across different degraded signal conditions. Additionally, our approach has demonstrated excellent homeostatic properties, maintaining stable neuronal firing rates under challenging conditions such as occlusions, low light levels, the presence of small or rapidly moving objects, and linear motion. Notably, it also exhibited the lowest power consumption among the models evaluated.

Our model effectively emulates certain biological neuron behaviours, but it does not fully capture all human neuron characteristics, highlighting an area for enhancement to improve biological authenticity. Future work could explore advanced dynamic threshold functions incorporating elements like dynamic refractory periods and network activity. Additionally, current neuromorphic hardware architectures are primarily restricted to Leaky Integrate-and-Fire (LIF) and Spike Response Model (SRM) configurations, which precludes the direct deployment of our algorithm on hardware directly. Addressing this limitation remains a pivotal area for future research to ensure robust real-world applicability.

% \newline
% \textbf{Acknowledgements}
\section*{Acknowledgements}
This work was supported by Kingston University, the Advanced Research and Innovation Center (ARIC), which is jointly funded by Khalifa University of Science and Technology, STRATA Manufacturing PJSC (a Mubadala company) and Sandooq Al Watan under Grant SWARD-S22-015. In addition, the authors would like to extend their appreciation to Ipsotek, an Eviden Company, for their generous support and valuable contributions to the research project.

\bibliographystyle{IEEEtran}

\bibliography{references}

@article{Lichtsteiner2008ASensor,
    title = {{A 128 × 128 120 dB 15 {$\mu$}s latency asynchronous temporal contrast vision sensor}},
    year = {2008},
    journal = {IEEE Journal of Solid-State Circuits},
    author = {Lichtsteiner, Patrick and Posch, Christoph and Delbruck, Tobi},
    number = {2},
    month = {2},
    pages = {566--576},
    volume = {43},
    doi = {10.1109/JSSC.2007.914337},
    issn = {00189200}
}

@article{Hao2018ARuleb,
    title = {{A Biologically Plausible Supervised Learning Method for Spiking Neural Networks Using the Symmetric STDP Rule}},
    year = {2018},
    author = {Hao, Yunzhe and Huang, Xuhui and Dong, Meng and Xu, Bo},
    month = {12},
    doi = {10.1016/j.neunet.2019.09.007},
    arxivId = {1812.06574}
}

@article{Li2023ANetworks,
    title = {{A Graph is Worth 1-bit Spikes: When Graph Contrastive Learning Meets Spiking Neural Networks}},
    year = {2023},
    author = {Li, Jintang and Zhang, Huizhe and Wu, Ruofan and Zhu, Zulun and Chen, Liang and Zheng, Zibin and Wang, Baokun and Meng, Changhua},
    month = {5},
    url = {http://arxiv.org/abs/2305.19306},
    arxivId = {2305.19306}
}

@article{Aydin2023APerception,
    title = {{A Hybrid ANN-SNN Architecture for Low-Power and Low-Latency Visual Perception}},
    year = {2023},
    author = {Aydin, Asude and Gehrig, Mathias and Gehrig, Daniel and Scaramuzza, Davide},
    month = {3},
    arxivId = {2303.14176}
}

@inproceedings{Amir2017ASystem,
    title = {{A Low Power, Fully Event-Based Gesture Recognition System}},
    year = {2017},
    booktitle = {2017 IEEE Conference on Computer Vision and Pattern Recognition (CVPR)},
    author = {Amir, Arnon and Taba, Brian and Berg, David and Melano, Timothy and McKinstry, Jeffrey and Di Nolfo, Carmelo and Nayak, Tapan and Andreopoulos, Alexander and Garreau, Guillaume and Mendoza, Marcela and Kusnitz, Jeff and Debole, Michael and Esser, Steve and Delbruck, Tobi and Flickner, Myron and Modha, Dharmendra},
    month = {7},
    pages = {7388--7397},
    publisher = {IEEE},
    isbn = {978-1-5386-0457-1},
    doi = {10.1109/CVPR.2017.781}
}

@article{huang2024neuromorphic,
  title={A neuromorphic dataset for tabletop object segmentation in indoor cluttered environment},
  author={Huang, Xiaoqian and Kachole, Sanket and Ayyad, Abdulla and Naeini, Fariborz Baghaei and Makris, Dimitrios and Zweiri, Yahya},
  journal={Scientific data},
  volume={11},
  number={1},
  pages={127},
  year={2024},
  publisher={Nature Publishing Group UK London}
}

@article{kachole2024bimodal,
  title={Bimodal SegNet: Fused instance segmentation using events and RGB frames},
  author={Kachole, Sanket and Huang, Xiaoqian and Naeini, Fariborz Baghaei and Muthusamy, Rajkumar and Makris, Dimitrios and Zweiri, Yahya},
  journal={Pattern Recognition},
  volume={149},
  pages={110215},
  year={2024},
  publisher={Elsevier}
}

@article{kachole2023bimodal,
  title={Bimodal SegNet: Instance segmentation fusing events and RGB frames for robotic grasping},
  author={Kachole, Sanket and Huang, Xiaoqian and Naeini, Fariborz Baghaei and Muthusamy, Rajkumar and Makris, Dimitrios and Zweiri, Yahya},
  journal={arXiv preprint arXiv:2303.11228},
  year={2023}
}

@article{Huang2023AEnvironment,
    title = {{A Neuromorphic Dataset for Object Segmentation in Indoor Cluttered Environment}},
    year = {2023},
    author = {Huang, Xiaoqian and Sanket, Kachole and Ayyad, Abdulla and Naeini, Fariborz Baghaei and Makris, Dimitrios and Zweiri, Yahya},
    month = {2},
    url = {http://arxiv.org/abs/2302.06301},
    arxivId = {2302.06301}
}

@article{Shaban2021AnImplementation.,
    title = {{An adaptive threshold neuron for recurrent spiking neural networks with nanodevice hardware implementation.}},
    year = {2021},
    journal = {Nature communications},
    author = {Shaban, Ahmed and Bezugam, Sai Sukruth and Suri, Manan},
    number = {1},
    month = {7},
    pages = {4234},
    volume = {12},
    doi = {10.1038/s41467-021-24427-8},
    issn = {2041-1723},
    pmid = {34244491}
}

@inproceedings{Cazalets2023AnCombination,
    title = {{An homeostatic activity-dependent structural plasticity algorithm for richer input combination}},
    year = {2023},
    booktitle = {2023 International Joint Conference on Neural Networks (IJCNN)},
    author = {Cazalets, Tanguy and Dambre, Joni},
    month = {6},
    pages = {1--8},
    publisher = {IEEE},
    url = {https://ieeexplore.ieee.org/document/10191230/},
    isbn = {978-1-6654-8867-9},
    doi = {10.1109/IJCNN54540.2023.10191230}
}

@article{Kim2022BeyondSegmentation,
    title = {{Beyond classification: directly training spiking neural networks for semantic segmentation}},
    year = {2022},
    journal = {Neuromorphic Computing and Engineering },
    author = {Kim, Youngeun and Chough, Joshua and Panda, Priyadarshini},
    number = {4},
    month = {12},
    pages = {44015},
    volume = {2},
    publisher = {IOP Publishing},
    url = {https://dx.doi.org/10.1088/2634-4386/ac9b86},
    doi = {10.1088/2634-4386/ac9b86}
}

@article{Ding2022BiologicallyNetworks,
    title = {{Biologically Inspired Dynamic Thresholds for Spiking Neural Networks}},
    year = {2022},
    journal = {Neural Information Processing Systems},
    author = {Ding, Jianchuan and Dong, Bo and Heide, Felix and Ding, Yufei and Zhou, Yunduo and Yin, Baocai and Yang, Xin}
}

@inproceedings{Bohnstingl2022Biologically-inspiredHardware,
    title = {{Biologically-inspired training of spiking recurrent neural networks with neuromorphic hardware}},
    year = {2022},
    booktitle = {2022 IEEE 4th International Conference on Artificial Intelligence Circuits and Systems (AICAS)},
    author = {Bohnstingl, Thomas and Surina, Anja and Fabre, Maxime and Demirag, Yigit and Frenkel, Charlotte and Payvand, Melika and Indiveri, Giacomo and Pantazi, Angeliki},
    month = {6},
    pages = {218--221},
    publisher = {IEEE},
    isbn = {978-1-6654-0996-4},
    doi = {10.1109/AICAS54282.2022.9869963}
}

@article{Li2017CIFAR10-DVS:Classification,
    title = {{CIFAR10-DVS: An Event-Stream Dataset for Object Classification}},
    year = {2017},
    journal = {Frontiers in Neuroscience},
    author = {Li, Hongmin and Liu, Hanchao and Ji, Xiangyang and Li, Guoqi and Shi, Luping},
    month = {5},
    volume = {11},
    doi = {10.3389/fnins.2017.00309},
    issn = {1662-453X}
}

@article{Orchard2015ConvertingSaccades,
    title = {{Converting Static Image Datasets to Spiking Neuromorphic Datasets Using Saccades}},
    year = {2015},
    journal = {Frontiers in Neuroscience},
    author = {Orchard, Garrick and Jayawant, Ajinkya and Cohen, Gregory K and Thakor, Nitish},
    volume = {9},
    url = {https://www.frontiersin.org/articles/10.3389/fnins.2015.00437},
    doi = {10.3389/fnins.2015.00437},
    issn = {1662-453X}
}

@article{OConnor2016DeepNetworks,
    title = {{Deep Spiking Networks}},
    year = {2016},
    journal = {arXiv},
    author = {O'Connor, Peter and Welling, Max},
    month = {2},
    arxivId = {1602.08323}
}

@inproceedings{Calabrese2019DHP19:Dataset,
    title = {{DHP19: Dynamic Vision Sensor 3D Human Pose Dataset}},
    year = {2019},
    booktitle = {2019 IEEE/CVF Conference on Computer Vision and Pattern Recognition Workshops (CVPRW)},
    author = {Calabrese, Enrico and Taverni, Gemma and Easthope, Christopher Awai and Skriabine, Sophie and Corradi, Federico and Longinotti, Luca and Eng, Kynan and Delbruck, Tobi},
    pages = {1695--1704},
    doi = {10.1109/CVPRW.2019.00217}
}

@article{Neftci2014Event-drivenSystems,
    title = {{Event-driven contrastive divergence for spiking neuromorphic systems}},
    year = {2014},
    journal = {Frontiers in Neuroscience},
    author = {Neftci, Emre and Das, Srinjoy and Pedroni, Bruno and Kreutz-Delgado, Kenneth and Cauwenberghs, Gert},
    volume = {7},
    doi = {10.3389/fnins.2013.00272},
    issn = {1662-453X}
}

@inproceedings{Abreu2023FlowHardware,
    title = {{Flow cytometry with event-based vision and spiking neuromorphic hardware}},
    year = {2023},
    booktitle = {2023 IEEE/CVF Conference on Computer Vision and Pattern Recognition Workshops (CVPRW)},
    author = {Abreu, Steven and Gouda, Muhammed and Lugnan, Alessio and Bienstman, Peter},
    month = {6},
    pages = {4139--4147},
    publisher = {IEEE},
    isbn = {979-8-3503-0249-3},
    doi = {10.1109/CVPRW59228.2023.00435}
}

@article{Makhlooghpour2016HighModel,
    title = {{High Accuracy Implementation of Adaptive Exponential Integrated and Fire Neuron Model}},
    year = {2016},
    journal = {IEEE World Congress on Computational Intelligence},
    author = {Makhlooghpour, Aliasghar and Soleimani, Hamid and Arash, Ahmadi and Mark, Zwolinski and Mehrdad, Saif},
    pages = {5301},
    volume = {},
    isbn = {9781509006205}
}

@article{Rao1999PredictiveEffects,
    title = {{Predictive coding in the visual cortex:  a functional interpretation of some extra-classical receptive-field effects}},
    year = {1999},
    journal = {Nature Neuroscience},
    author = {Rao, Rajesh P N and Ballard, Dana H},
    number = {1},
    pages = {79--87},
    volume = {2},
    url = {https://doi.org/10.1038/4580},
    doi = {10.1038/4580},
    issn = {1546-1726}
}

@article{Kim2022RateNetworks,
    title = {{Rate Coding or Direct Coding: Which One is Better for Accurate, Robust, and Energy-efficient Spiking Neural Networks?}},
    year = {2022},
    journal = {ICASSP, IEEE International Conference on Acoustics, Speech and Signal Processing - Proceedings},
    author = {Kim, Youngeun and Park, Hyoungseob and Moitra, Abhishek and Bhattacharjee, Abhiroop and Venkatesha, Yeshwanth and Panda, Priyadarshini},
    pages = {71--75},
    volume = {2022-May},
    publisher = {Institute of Electrical and Electronics Engineers Inc.},
    isbn = {9781665405409},
    doi = {10.1109/ICASSP43922.2022.9747906},
    issn = {15206149},
    arxivId = {2202.03133}
}

@techreport{WangRegulationModel,
    title = {{Regulation of Spontaneous Rhythmic Activity and Preserved Stimulus Dependent Pattern by STDP in the Hippocampal CA3 Model}},
    author = {Wang, Lip0 and Rajapakse, Jagath C and Fukushima, Kunihiko and Lee, Soo-Young and Yao, Xin and Yoshida', Motoharu and Hayashi2, Hatsuo},
    volume = {1}
}

@article{Wu2018Spatio-TemporalNetworks,
    title = {{Spatio-Temporal Backpropagation for Training High-Performance Spiking Neural Networks}},
    year = {2018},
    journal = {Frontiers in Neuroscience},
    author = {Wu, Yujie and Deng, Lei and Li, Guoqi and Zhu, Jun and Shi, Luping},
    month = {5},
    volume = {12},
    doi = {10.3389/fnins.2018.00331},
    issn = {1662-453X}
}

@article{Fontaine2014Spike-thresholdVivo.,
    title = {{Spike-threshold adaptation predicted by membrane potential dynamics in vivo.}},
    year = {2014},
    journal = {PLoS computational biology},
    author = {Fontaine, Bertrand and Pe{\~{n}}a, José Luis and Brette, Romain},
    number = {4},
    month = {4},
    pages = {e1003560},
    volume = {10},
    doi = {10.1371/journal.pcbi.1003560},
    issn = {1553-7358},
    pmid = {24722397}
}

@article{Kim2021SpikingUpdate.,
    title = {{Spiking Neural Network (SNN) With Memristor Synapses Having Non-linear Weight Update.}},
    year = {2021},
    journal = {Frontiers in computational neuroscience},
    author = {Kim, Taeyoon and Hu, Suman and Kim, Jaewook and Kwak, Joon Young and Park, Jongkil and Lee, Suyoun and Kim, Inho and Park, Jong-Keuk and Jeong, YeonJoo},
    pages = {646125},
    volume = {15},
    doi = {10.3389/fncom.2021.646125},
    issn = {1662-5188},
    pmid = {33776676}
}

@inproceedings{Lehmann2021SpikingBiCMOS,
    title = {{Spiking Neural Networks based Rate-Coded Logic Gates for Automotive Applications in BiCMOS}},
    year = {2021},
    booktitle = {2021 IEEE International Conference on Microwaves, Antennas, Communications and Electronic Systems, COMCAS 2021},
    author = {Lehmann, Hendrik M. and Hille, Julian and Grassmann, Cyprian and Issakov, Vadim},
    pages = {280--285},
    publisher = {Institute of Electrical and Electronics Engineers Inc.},
    isbn = {9780738146720},
    doi = {10.1109/COMCAS52219.2021.9629011}
}

@book{Gerstner2002SpikingModels,
    title = {{Spiking Neuron Models}},
    year = {2002},
    author = {Gerstner, Wulfram and Kistler, Werner M.},
    month = {8},
    publisher = {Cambridge University Press},
    address = {Cambridge},
    isbn = {9780521813846},
    doi = {10.1017/CBO9780511815706}
}

@article{Gerstner2002SpikingPlasticity,
    title = {{Spiking Neuron Models: Single Neurons, Populations, Plasticity}},
    year = {2002},
    author = {Gerstner, Wulfram and Kistler, Werner M.},
    month = {8},
    publisher = {Cambridge University Press},
    isbn = {9780521813846},
    doi = {10.1017/CBO9780511815706}
}

@inproceedings{Deng2021STDPClassification,
    title = {{STDP and Competition Learning in Spiking Neural Networks and its application to Image Classification}},
    year = {2021},
    booktitle = {2021 International Conference on Information, Cybernetics, and Computational Social Systems, ICCSS 2021},
    author = {Deng, Min and Li, Chuandong},
    pages = {385--389},
    publisher = {Institute of Electrical and Electronics Engineers Inc.},
    isbn = {9781665402453},
    doi = {10.1109/ICCSS53909.2021.9722029}
}

@article{Yeung2004SynapticModel,
    title = {{Synaptic homeostasis and input selectivity follow from a calcium-dependent plasticity model}},
    year = {2004},
    journal = {Proceedings of the National Academy of Sciences},
    author = {Yeung, Luk Chong and Shouval, Harel Z. and Blais, Brian S. and Cooper, Leon N.},
    number = {41},
    month = {10},
    pages = {14943--14948},
    volume = {101},
    doi = {10.1073/pnas.0405555101},
    issn = {0027-8424}
}

@article{Gerstner1995TimeModels.,
    title = {{Time structure of the activity in neural network models.}},
    year = {1995},
    journal = {Physical review. E, Statistical physics, plasmas, fluids, and related interdisciplinary topics},
    author = {Gerstner, W},
    number = {1},
    month = {1},
    pages = {738--758},
    volume = {51},
    doi = {10.1103/physreve.51.738},
    issn = {1063-651X},
    pmid = {9962697}
}

@article{Baldwin2023Time-OrderedCameras,
    title = {{Time-Ordered Recent Event (TORE) Volumes for Event Cameras}},
    year = {2023},
    journal = {IEEE Transactions on Pattern Analysis and Machine Intelligence},
    author = {Baldwin, R. Wes and Liu, Ruixu and Almatrafi, Mohammed and Asari, Vijayan and Hirakawa, Keigo},
    number = {2},
    month = {2},
    pages = {2519--2532},
    volume = {45},
    doi = {10.1109/TPAMI.2022.3172212},
    issn = {0162-8828}
}

@article{Pozo2010UnravelingPlasticity,
    title = {{Unraveling Mechanisms of Homeostatic Synaptic Plasticity}},
    year = {2010},
    journal = {Neuron},
    author = {Pozo, Karine and Goda, Yukiko},
    number = {3},
    month = {5},
    pages = {337--351},
    volume = {66},
    doi = {10.1016/j.neuron.2010.04.028},
    issn = {08966273}
}

@article{Ferre2018UnsupervisedSTDP,
    title = {{Unsupervised Feature Learning With Winner-Takes-All Based STDP}},
    year = {2018},
    journal = {Frontiers in Computational Neuroscience},
    author = {Ferr{\'{e}}, Paul and Mamalet, Franck and Thorpe, Simon J.},
    month = {4},
    volume = {12},
    doi = {10.3389/fncom.2018.00024},
    issn = {1662-5188}
}

@article{leng2018spiking,
  title={Spiking neurons with short-term synaptic plasticity form superior generative networks},
  author={Leng, Luziwei and Martel, Roman and Breitwieser, Oliver and Bytschok, Ilja and Senn, Walter and Schemmel, Johannes and Meier, Karlheinz and Petrovici, Mihai A},
  journal={Scientific reports},
  volume={8},
  number={1},
  pages={10651},
  year={2018},
  publisher={Nature Publishing Group UK London}
}

@article{bellec2018long,
  title={Long short-term memory and learning-to-learn in networks of spiking neurons},
  author={Bellec, Guillaume and Salaj, Darjan and Subramoney, Anand and Legenstein, Robert and Maass, Wolfgang},
  journal={Advances in neural information processing systems},
  volume={31},
  year={2018}
}

@article{bellec2020solution,
  title={A solution to the learning dilemma for recurrent networks of spiking neurons},
  author={Bellec, Guillaume and Scherr, Franz and Subramoney, Anand and Hajek, Elias and Salaj, Darjan and Legenstein, Robert and Maass, Wolfgang},
  journal={Nature communications},
  volume={11},
  number={1},
  pages={3625},
  year={2020},
  publisher={Nature Publishing Group UK London}
}

@article{zhang2023accurate,
  title={Accurate and Efficient Event-based Semantic Segmentation Using Adaptive Spiking Encoder-Decoder Network},
  author={Zhang, Rui and Leng, Luziwei and Che, Kaiwei and Zhang, Hu and Cheng, Jie and Guo, Qinghai and Liao, Jiangxing and Cheng, Ran},
  journal={arXiv preprint arXiv:2304.11857},
  year={2023}
}

@article{zhang2023automotive,
  title={Automotive Object Detection via Learning Sparse Events by Temporal Dynamics of Spiking Neurons},
  author={Zhang, Hu and Leng, Luziwei and Che, Kaiwei and Liu, Qian and Cheng, Jie and Guo, Qinghai and Liao, Jiangxing and Cheng, Ran},
  journal={arXiv preprint arXiv:2307.12900},
  year={2023}
}

@inproceedings{kim2021n,
  title={N-imagenet: Towards robust, fine-grained object recognition with event cameras},
  author={Kim, Junho and Bae, Jaehyeok and Park, Gangin and Zhang, Dongsu and Kim, Young Min},
  booktitle={Proceedings of the IEEE/CVF international conference on computer vision},
  pages={2146--2156},
  year={2021}
}

@article{sanes1993activity,
  title={Activity-dependent refinement of inhibitory connections},
  author={Sanes, Dan H and Tak{\'a}cs, Catherine},
  journal={European Journal of Neuroscience},
  volume={5},
  number={6},
  pages={570--574},
  year={1993},
  publisher={Wiley Online Library}
}

@article{lu2003bdnf,
  title={BDNF and activity-dependent synaptic modulation},
  author={Lu, Bai},
  journal={Learning \& memory},
  volume={10},
  number={2},
  pages={86--98},
  year={2003},
  publisher={Cold Spring Harbor Lab}
}

@article{yu2017energy,
  title={Energy-efficient neural information processing in individual neurons and neuronal networks},
  author={Yu, Lianchun and Yu, Yuguo},
  journal={Journal of Neuroscience Research},
  volume={95},
  number={11},
  pages={2253--2266},
  year={2017},
  publisher={Wiley Online Library}
}

@article{sanket20163Tracker,
    title = {{3 Dimensional Welding SPM/Path Tracker}},
    year = {2016},
    journal = {International Journal Of Design And Manufacturing Technology},
    author = {sanket, Kachole and Manish, Mahakal and Anurag, Bhagwatkar},
    number = {3},
    month = {12},
    volume = {7},
    publisher = {IAEME Publication Chennai},
    doi = {10.34218/ijdmt.7.3.2016.003},
    issn = {0976-6995}
}

@article{Kachole2020AHoneybees,
    title = {{A Computer Vision Approach to Monitoring the Activity and Well-Being of Honeybees}},
    year = {2020},
    journal = {Intelligent Environments},
    author = {Kachole, Sanket and Hunter, Gordon and Duran, Olga},
    url = {https://www.axis.com/en-gb/products/axis-p1346},
    doi = {10.3233/AISE200036}
}

@article{naeini2022event,
  title={Event augmentation for contact force measurements},
  author={Naeini, Fariborz Baghaei and Kachole, Sanket and Muthusamy, Rajkumar and Makris, Dimitrios and Zweiri, Yahya},
  journal={IEEE Access},
  volume={10},
  pages={123651--123660},
  year={2022},
  publisher={IEEE}
}

@inproceedings{kachole2024asynchronous,
  title={Asynchronous bioplausible neuron for spiking neural networks for event-based vision},
  author={Kachole, Sanket and Sajwani, Hussain and Naeini, Fariborz Baghaei and Makris, Dimitrios and Zweiri, Yahya},
  booktitle={European Conference on Computer Vision},
  pages={399--415},
  year={2024},
  organization={Springer}
}

\end{document}